\documentclass[letterpaper, 10 pt, conference]{ieeeconf} 

\IEEEoverridecommandlockouts
\usepackage{cite}
\usepackage{amsmath,amssymb,amsfonts}
\usepackage{algorithmic}
\usepackage{graphicx}
\usepackage{footnote}
\usepackage{textcomp}
\usepackage{xcolor}
\usepackage{comment}
\usepackage{lipsum}
\usepackage{algorithm}
\usepackage{url}
\usepackage{svg}

\usepackage{titlesec}
\titlespacing*{\section}{0pt}{0.3\baselineskip}{0.2\baselineskip}

\usepackage{changepage}
\newcommand{\todo}[1]{}
\newcommand{\etal}[1]{#1 et al.}
\def\BibTeX{{\rm B\kern-.05em{\sc i\kern-.025em b}\kern-.08em
    T\kern-.1667em\lower.7ex\hbox{E}\kern-.125emX}}
\IEEEoverridecommandlockouts 
\begin{document}

\title{\LARGE \bf
MDT3D: Multi-Dataset Training for LiDAR\\
3D Object Detection Generalization\\

}

\author{Louis Soum-Fontez${^*}$\thanks{${^*}$Centre for Robotics, Mines Paris - PSL, PSL University, 75006 Paris, France {firstname.surname@minesparis.psl.eu}}, Jean-Emmanuel Deschaud$^*$, François Goulette${^*}{^\dagger}$\thanks{$^\dagger$U2IS, ENSTA Paris, Institut Polytechnique de Paris, 91120 Palaiseau, France}}

\maketitle

\begin{abstract}
Supervised 3D Object Detection models have been displaying increasingly better performance in single-domain cases where the training data comes from the same environment and sensor as the testing data. 
However, in real-world scenarios data from the target domain may not be available for finetuning or for domain adaptation methods. Indeed, 3D object detection models trained on a source dataset with a specific point distribution have shown difficulties in generalizing to unseen datasets. 
Therefore, we decided to leverage the information available from several annotated source datasets with our Multi-Dataset Training for 3D Object Detection (MDT3D) method to increase the robustness of 3D object detection models when tested in a new environment with a different sensor configuration. To tackle the labelling gap between datasets, we used a new label mapping based on coarse labels.
Furthermore, we show how we managed the mix of datasets during training and finally introduce a new cross-dataset augmentation method: cross-dataset object injection. We demonstrate that this training paradigm shows improvements for different types of 3D object detection models. The source code and additional results for this research project will be publicly available on GitHub for interested parties to access and utilize: 

\url{https://github.com/LouisSF/MDT3D}

\end{abstract}

\section{Introduction}

The emergence of big data and deep learning, along with the availability and decreasing costs of 3D LiDAR sensors for autonomous vehicles, have led to the emergence of new approaches and datasets for detecting objects in 3D, thus becoming a core part of the autonomous vehicle technology stack. Indeed, these sensors allow for more precise depth estimation than cameras. In the case of autonomous vehicles, the task of 3D object detection is to detect road users as 3D bounding boxes. Methods that use deep learning for 3D object detection with LiDAR have made great strides in the past decade. 
However, 3D object detection models suffer from poor generalization ability due to the wide range of sensor configurations available, more so than their 2D counterparts which generally use cameras that output similar types of RGB images for a given scene. A model performing well on one dataset may break down when transferred to a new city or sensor. Some studies focus on closing the gap between a source dataset and a target deployment dataset, a task called \textit{domain adaptation}. However, these methods rely on the availability of target data for iteratively refining a trained model, which is a strong constraint for real-world scenarios. 
We believe that focusing on the training data instead of the detection model could be the key to a better overall detection method.
The vast majority of 3D object detection papers have focused on a single dataset for training. Indeed, this simplifies the training process and is the preferred method for highlighting a new architecture. However, in the specific case of domain generalization, we believe that models tend to overfit on the geometric specifications of the dataset used.

In this paper, we propose a new training paradigm for 3D object detection that uses cross-dataset variance to robustify detection models. Two apparent roadblocks for this paradigm are the unequal dataset sizes and the different labels sets used for each dataset. To circumvent these issues, we introduce a coarse label set that can be used in any LiDAR dataset for autonomous driving. Using this label set allowed us to optimize our models with homogenized labels. Furthermore, we leveraged our multi-dataset training paradigm with a novel cross-dataset augmentation to decrease the reliance on dataset-specific context around objects.

We summarize our contributions as follows:

\begin{itemize}
    \item Analyzing the limitations of model transfer on new unseen datasets from single-dataset training.
    \item Introducing a cross-dataset training paradigm with a common label set for unifying classes of semantic similarity across datasets.
    \item Using this training paradigm and our new cross-dataset augmentation to show improved generalization ability of the trained models across several unseen datasets.
\end{itemize}

\begin{figure*}[h!]
\vspace{4pt}
\centerline{\includegraphics[width=1\textwidth]{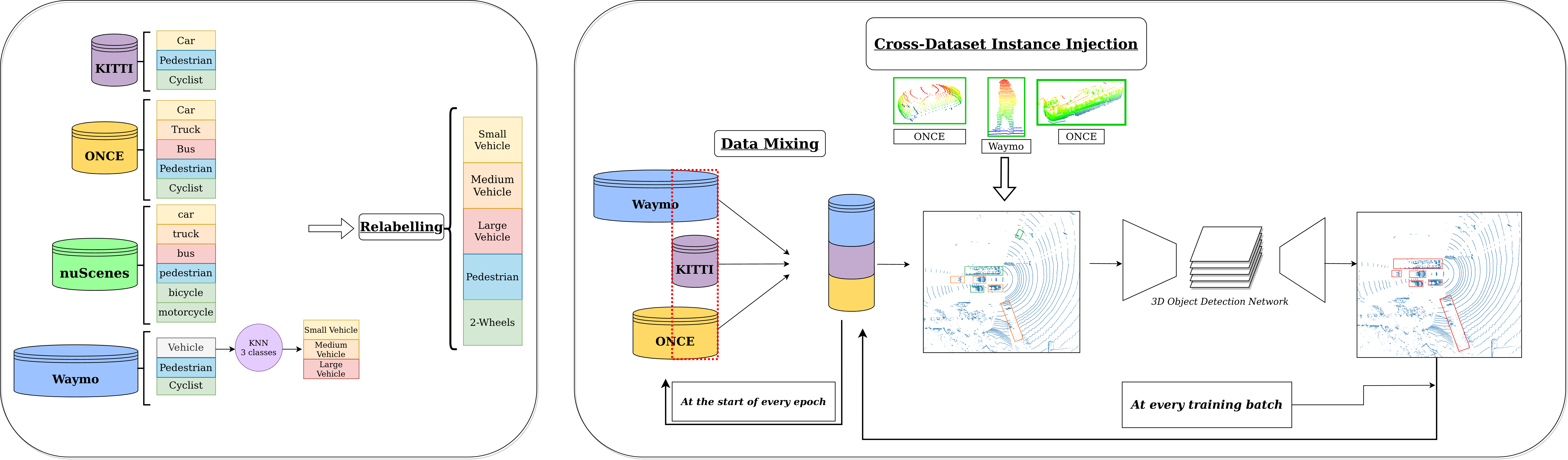}}

\caption{\textbf{M}ulti-\textbf{D}ataset \textbf{T}raining for \textbf{3D} object detection (MDT3D) Pipeline. We preprocess the label sets of our training datasets to associate them to a joint label set. At training time, we sample scans from each dataset at the beginning of each training epochs. We then augment input scans using cross-dataset injections of object instances, before using the final 3D scan as input to a 3D object detection model. This figure represents the training procedure when nuScenes is the target dataset}
\label{fig:pipe}
\end{figure*}

\section{Related Works}

\subsection{3D Object Detection} 

While Convolutional Neural Networks (CNNs) are considered state of the art in 2D object detection, their application in 3D is more complex due to the unordered nature of point clouds. \etal{Qi}\cite{qi2016pointnet} aimed to resolve this by proposing an architecture that operates on individual points and applies a maxpooling operation on local point neighborhoods.
Many methods take advantage of the local shape features extracted by PointNet by subdividing the 3D space and applying a PointNet module on each voxel\cite{Zhou_Tuzel_2018}, followed by 2D or 3D convolutional filters.
 
While voxel-based methods are predominant, point-based methods also exist. PointRCNN\cite{Shi_2019_CVPR}
subdivides the input point cloud into three according to their distance to the sensor then applies a PointNet++\cite{Qi2017PointNetDH} on each subdivision to obtain multi-range features.

PV-RCNN~\cite{Shi_2020_CVPR} combines point and voxel features at several steps, refining initial proposals using features associated to a set of keypoints. CenterPoint~\cite{Yin_2021_CVPR} was inspired by recent 2D detection methods~\cite{zhou2019objects} that aim to predict a heatmap of object center locations, extract object centers from local peaks of the heatmap, and directly regress object sizes. 

\begin{table*}[t]
\vspace{5pt}
\caption{\label{tab:lidar_stat}Characteristics of 4 datasets in autonomous driving containing LiDAR data with object annotations}
\begin{center}
\scalebox{0.85}{%
\begin{tabular}{|c||c|c|c|c|c|c|c|c|c|c|c|}
 \hline
  \cline{1-11}
 & \multicolumn{2}{|c|}{\textbf{Dataset Info}} & \multicolumn{3}{|c|}{\textbf{Sensor Characteristics}} & \multicolumn{5}{|c|}{\textbf{Avg Number of Points Contained in Bounding Boxes}} \\
 \cline{1-11}
 \textbf{\textit{Name}} & \textbf{\textit{Annotated frames}} & \textbf{\textit{3D boxes}} & \textbf{\textit{LiDAR type}} & \textbf{\textit{VFOV}} & \textbf{\textit{Average points per beam}} & \textbf{\textit{Car}} & \textbf{\textit{Bus}} & \textbf{\textit{Truck}} & \textbf{\textit{Pedestrian}} & \textbf{\textit{Cyclist}}\\
 \hline
 KITTI  & 15K & 200K & 64-beam & [-23.6°, 3.2°] & 1863 &  253 & - & - & 155 & 135 \\
  \hline
 ONCE & 16K  & 417K & 40-beam & [-25°, 15°] & 1593 & 424 & 1962 & 941 & 51 & 72\\
 \hline
 nuScenes  & 40K & 1.4M & 32-beam & [-30.0°, 10.0°] & 1084 & 108 & 291 & 184 & 12 & 17\\
  \hline
 Waymo & 200K & 15M & 64-beam & [-17.6°, 2,4°] & 2258 & 530
 & - & - & 91 & 163\\

 \hline
\end{tabular}}
\end{center}
\end{table*}

\subsection{Domain Gap Analysis}

\etal{Wang}~\cite{Wang_2020_CVPR} were the first to analyze the variations of performance of a given 3D object detector across datasets and found that the diversity of object sizes between geographical regions was a predominant factor of degradation in performance. To reduce such performance gaps, they suggested using a statistical normalization methods based on the dimensions of cars in the target domain. However, this necessitates statistics on the annotations of the target domain, which are rarely available in real-world scenarios. We add to their results by measuring transfer performance for several configurations of source and target datasets in Section \ref{sec:prelim}. While in 2D object detection, domain-adversarial methods have been widespread for domain adaptation~\cite{da_by_backprop}
, the disparities between LiDAR sensors have led to less success for these methods in 3D object detection.

Therefore, many researchers have analyzed and tackled the sensor-to-sensor gap. \etal{Richter}\cite{understanding_sensor2sensor} showed that 3D object detectors suffer from a strong dependence on the number of LiDAR points per object. 

\etal{Théodose}\cite{lidar_res_agnostic} generated a diversity of point cloud densities using data augmentation by randomly dropping some points from the same LiDAR beam layer.
SPG~\cite{Xu_2021_ICCV} generates points to complete missing foreground regions. However, the authors assumed that the same LiDAR sensor is used, which leads to identical LiDAR pattern distributions, and that rays are simply dropped due to occlusion, weather interferences, or low reflectance. 

\subsection{Domain Generalization}

Domain generalization is the task of training on one or several source domains and evaluating them on unknown target domains. This has been a difficult challenge for 2D object detection methods due to the high variety of environments and classes for each dataset. \etal{Zhang}\cite{Zhang2022TowardsDG} formulated a comprehensive evaluation benchmark and proposed a method of eliminating the dependence within RoI features, thus improving the performance of existing detectors in domain generalization tasks.
For 2D object detection in autonomous vehicle settings, \etal{Simonelli}\cite{10.1007/978-3-030-58542-6_46} used virtual views that normalize the appearance of objects according to their distance from the camera.
The task of 3D object detection domain generalization remains a largely unexplored topic, researchers have investigated it such as \etal{Lehner}\cite{lehner2022_3d-vfield}, who used a data augmentation method that deforms point clouds via vector fields learned in an adversarial fashion during training. However, they only evaluated the \textit{Car} class and focused on the sensor-to-sensor domain gap.  

\vspace{-3pt}
\subsection{Multi-Dataset Training}

To increase the generalization of different detection models, selections of the training procedure and datasets is of utmost importance. One way of merging several types of scenes and configurations is to use multiple datasets for training. In the 2D realm, \etal{Zhao}\cite{unifiedlabelspace} identified the inconsistency of annotations between different datasets. They trained dataset-specific detectors, used them to re-annotate each dataset, and then used them to train a single detector using all labels. However, they mainly considered the case in which datasets have non-intersecting label sets but common classes have similar representations.
Simple Multi-Dataset Detection\cite{simplemultidsdet} reuses partitioned dataset-specific detectors, but the label space becomes a part of the training procedure, and the mapping from dataset-specific labels to a unified label space is optimized to minimize performance loss across individual dataset detectors.
In the case of point clouds, \etal{Sanchez}~\cite{CoLA} used intermediate-level coarse labels to pretrain their 3D point cloud segmentation models using several datasets.

To our knowledge, we are the first to use multiple datasets at once for training a 3D object detection model for the generalization task.

\section{Domain Gap Analysis}\label{sec:prelim}

\begin{table*}[t]
  \caption{\label{tab:class_stat}Class Statistics per Dataset. "-" means that the dataset does not contain that class.}
  \begin{center}
  \scalebox{0.9}{
    \begin{tabular}{|c||c|c|c|c|c|c|c|c|c|c|}
    \hline
    \cline{1-11}
     & \multicolumn{5}{|c|}{\textbf{Average Size $(l\times w\times h) (m)$}} & \multicolumn{5}{|c|}{\textbf{Average Number of Class Instances per Scan}} \\
     \cline{1-11}
    \textbf{\textit{Name}} & \textbf{\textit{Car}} & \textbf{\textit{Bus}} & \textbf{\textit{Truck}} & \textbf{\textit{Pedestrian}} & \textbf{\textit{Cyclist}} & \textbf{\textit{Car}} & \textbf{\textit{Bus}} & \textbf{\textit{Truck}} & \textbf{\textit{Pedestrian}} & \textbf{\textit{Cyclist}} \\
    \hline
    KITTI & $3.9\times1.6\times1.5$  & - & - & $ 0.8\times0.6\times1.8 $ & $ 1.8\times0.6\times1.7$ &  3.8  & - & - & 0.6 & 0.2\\
    \hline
    ONCE & $ 4.4\times1.8\times1.6$ & $10.7\times2.9\times3.3$ & $6.4\times2.4\times2.5$ & $ 0.8\times0.8\times1.7 $ & $2.1\times0.8\times1.3$ & 19.8 & 0.6 & 0.6 & 2.9 & 6.3\\
    \hline
    nuScenes & $4.6\times2.0\times1.7$ & $10.7\times2.9\times3.4$ & $7.3\times2.5\times3.0$ & $0.7\times0.7\times1.8$ & $1.7\times0.6\times1.3$ & 11 & 0.5 & 2.2 & 5.7 & 0.3\\
    \hline
    Waymo & $ 4.6\times2.1\times1.7$ & - & -& $0.9\times0.8\times1.7$ & $1.7\times0.8\times1.7$ & 29.8 & - & - & 13.5 & 0.3\\
    \hline
    \end{tabular}
    }
   \end{center}
\end{table*}

In this section, we present the domain gaps between some common outdoor LiDAR 3D object detection datasets: KITTI\cite{Geiger2012CVPR}, nuScenes\cite{caesar2020nuscenes}, Waymo\cite{Sun_2020_CVPR}, and ONCE\cite{once_ds}. Details on the numbers of LiDAR scans and 3D bounding boxes are shown in Table \ref{tab:lidar_stat}.

\subsection{Sources of Domain Gaps}

Unlike RGB images, in which the representation of data is generally consistent but the domain gap comes from differences in object distributions or environments, 3D point clouds processed from LiDAR sensors differ in the point distributions due to sensor-specific characteristics. 
However, some other factors of domain gaps in images, such as the \textit{day-to-night} scenario, become negligible with LiDAR data.

Four main sources of domain gaps are as follows:
\begin{itemize}
    \item \textbf{Sensor-to-sensor}: Factors of differences in sensor characteristics, such as the range, the resolution (i.e, the number of beams), the vertical field of view,  reflectivity, or noise estimates (see Table \ref{tab:lidar_stat}). 
    These can impact the surface representations of objects as well as the level of detail per object.
    \item \textbf{Geography-to-geography}: Differences in environmental regions and scene contexts. This factor also impacts the size of objects, such as differences in car sizes between European and American regions.
    \item \textbf{Weather-to-weather}: Gaps between seasonal and weather scenarios. For instance, fog may result in numerous LiDAR rays having a weak return or being dropped. 
    \item \textbf{Class distributions}: Gaps in the distributions and annotations of the classes of different datasets. Some novel classes may be present in an unseen dataset, or manual annotations may classify objects differently. For instance, cars and trucks are considered two different classes in the ONCE dataset, whereas the Waymo Open Dataset contains only one vehicle class that does not distinguish the two.
\end{itemize}

\subsection{Statistical Analysis}

To find gaps between the datasets that could explain performance differences, we proceed to examine the dataset statistics. First, we isolated objects by class and computed their average bounding box sizes, in a $l \times w \times h$ format, where $l$, $w$, and $h$ are the length, width and height of the bounding boxes respectively. The results are shown in Table \ref{tab:class_stat}. We found significant differences between the datasets. These might come from the differences in annotation methods: for example the cyclists class considered only the actual bicycle in the case of ONCE but also considered the rider in the case of KITTI, which explains the height difference. On the other hand, some other gaps might have come from the actual differences in the object sizes between regions, such as differences in volumes between the German cars from the KITTI dataset and the American cars of the Waymo dataset. These disparities must be considered, as they result in the computation of significantly different regression targets for 3D detectors.

We filtered objects according to their class and computed both the average number of points in the bounding boxes of this class (see Table \ref{tab:lidar_stat}) and the average number of objects of this class per scene (see Table \ref{tab:class_stat}). We noticed a large gap both in the number of objects depending on their class and in the number of points within bounding boxes. While the Waymo dataset does not contain fine-grained class information about its vehicles, its high number of points on vehicles can be a strong tool for learning representations of this class. Furthermore, a lower-resolution LiDAR does not necessarily mean there is a lower number of points. For instance, cars in the ONCE dataset are generally captured in a setting of tight roads in dense Chinese cities, which means that they are generally closer to the sensor than are the cars in the KITTI dataset, which are generally captured in a setting of wider roads and in a suburban German setting. Due to the high number of certain objects, such as cyclists, in the ONCE dataset, models will easily overfit on this class, and we can expect massive degradation of performance on this same class when predicting bounding boxes on the nuScenes dataset due to its low number of cyclists and an average number of points that is five times lower than that in its ONCE equivalent.

\etal{Mao}\cite{once_ds} underlined differences in geography and time of day between the datasets. The time setting had a greater influence their 2D works than in ours, though it can affect the object density of scenes. Most importantly, the different geographical regions highlight a gap that introduces large contextual biases in each dataset, such as the width of the road, the building densities or the urban layouts. 

\subsection{Performance Degradation with Direct Transfer}

Some studies have described difficulties with model transfers, such as in ST3D\cite{Yang_2021_CVPR}, but have mainly focused on only one or two datasets. We expand this analysis comparing model performances on both CenterPoint\cite{Yin_2021_CVPR} and PointRCNN\cite{Shi_2019_CVPR} when transferring from a model trained on one dataset to a new unseen domain to underline the influence of domain gaps on 3D detectors, whether they are voxel-based or point-based. Our results are shown in Table \ref{tab:prelim_transfer}. The Oracle case, in which the training and testing datasets were the same, is also included as an upper bound. 

We also characterize each model using the mean of the transfer accuracies and the range of the accuracies for a given target dataset. The evaluation metric used is the mean average precision (mAP), which detailed in Section \ref{sec:setup}. In addition, we use the coarse labels that we define in Section \ref{sec:coarse_labels} to allow for a fair comparison, as we need a way to map labels between datasets.

We found that models tend to generalize quite poorly and that in terms of generalization, for a given configuration, not all datasets are created equal. This is because configurations in which only front-view objects are annotated like in KITTI, as well as short vehicle models trained on KITTI, tend to have poor accuracy on other datasets, such as on ONCE ($-50.3$ mAP compared to the Oracle model on CenterPoint and $-25.4$ mAP for PointRCNN).

For further analysis, Table \ref{tab:prelim_best_source} shows the best source dataset to use for transferring to a target domain for different classes when using a CenterPoint model. These results highlight the difficulty of choosing a single source dataset to use when transferring to a unknown dataset. We also include detailed metrics for individual classes in Table \ref{tab:res1_1}.

A general conclusion is that models trained on one dataset tend to have good accuracy on one or two classes in the target domain but break down with some others. One example is the use of the Waymo dataset for training, which tends to lead to a strong generalization ability for pedestrians but a poor one for vehicles when there is a strong difference in dimensions.

\begin{table}[h]
\vspace{7pt}
\caption[]{\label{tab:prelim_transfer}mAP results when transferring from a single dataset.\footnotemark}
\centering
\scalebox{0.75}{
\begin{tabular}{|c||c|c|c|c|c|c|c|c|}
\hline
\multicolumn{5}{|c|}{\textbf{CenterPoint}} \\
\hline
\hline
& \multicolumn{4}{|c|}{\textbf{\textit{Target Dataset}}} \\
\hline
\textbf{\textit{Training Dataset}}& \multicolumn{1}{|c|}{\textbf{\textit{KITTI}}} & \multicolumn{1}{|c|}{\textbf{\textit{ONCE}}} & \multicolumn{1}{|c|}{\textbf{\textit{nuScenes}}} & \multicolumn{1}{|c|}{\textbf{\textit{Waymo}}} \\
 \hline
 KITTI Only & 47.4$^*$ & 9.9 & 1.6& 3.0 \\
ONCE Only & 43.0 & 60.2$^*$ &  5.8& 13.0 \\
nuScenes Only & 12.6 & 10.7& 18.4$^*$& 0.9\\
Waymo Only & 34.2 & 28.8 & 7.2& 40.2$^*$ \\
\hline
mean (min-max) & 29.9 (12.6-43.0) & 16.5 (9.9-28.8) & 4.9 (1.6-7.2) & 5.6 (0.9-13.0) \\
\hline

\noalign{\vskip 4mm} 

\hline
\multicolumn{5}{|c|}{\textbf{PointRCNN}} \\
\hline
\hline
& \multicolumn{4}{|c|}{\textbf{\textit{Target Dataset}}} \\
\hline
\textbf{\textit{Training Dataset}} & \multicolumn{1}{|c|}{\textbf{\textit{KITTI}}} & \multicolumn{1}{|c|}{\textbf{\textit{ONCE}}} & \multicolumn{1}{|c|}{\textbf{\textit{nuScenes}}} & \multicolumn{1}{|c|}{\textbf{\textit{Waymo}}} \\
\hline
KITTI Only & 60.6$^*$ & 10.9 &  4.8& 3.0 \\
ONCE Only & 48.9 & 36.3$^*$&  7.9 & 7.8 \\
nuScenes Only & 23.0 & 10.2& 10.5$^*$& 5.9\\
Waymo Only & 32.3 & 16.0 & 7.6& 20.3$^*$ \\
\hline
mean (min-max) &  34.7 (23.0-48.9) & 12.4 (10.2-16.0) & 6.8 (4.8-7.9)&5.6 (3.0-7.8) \\
\hline
\end{tabular}}

\end{table}

\footnotetext{mAP computations on KITTI as a target set do not take into account the unavailable "Medium Vehicle" and "Large Vehicle" classes. "*" represents the case where the source and target datasets are the same, which we do not take into account when computing the mean and accuracy range.}

\begin{table}[h]
\caption[]{\label{tab:prelim_best_source}Best single source training dataset to use for a given target dataset and class, using a CenterPoint model.\footnotemark }
\centering
\scalebox{0.8}{

\begin{tabular}{|c|c|c|c|c|c|}
\hline
& \multicolumn{5}{|c|}{\textbf{\textit{Class}}} \\
\hline

\textbf{\textit{Target Dataset}} & \textbf{\textit{SmallV}} & \textbf{\textit{MedV}} & \textbf{\textit{LargeV}} & \textbf{\textit{2-Wheels}} & \textbf{\textit{Ped}} \\
\hline
KITTI & ONCE & - & -& Waymo& Waymo\\
ONCE & Waymo & nuScenes & Waymo& Waymo& Waymo\\
nuScenes & Waymo & Waymo & ONCE& ONCE& Waymo\\
Waymo & ONCE& ONCE& ONCE& ONCE& ONCE\\
\hline
\end{tabular}}

\end{table}

\footnotetext{Classes not present in the target dataset are noted as "-"}

\section{Multi-Dataset Training}

Based on our observations in Section \ref{sec:prelim}, we found that models trained on a single dataset perform poorly when directly transferred to a new dataset without additional training or hyperparameter tuning, particularly compared to an oracle model trained on the target dataset. Estimating the impact of model transfer necessitates information on both the source and target domains, but the latter is considered unavailable for the task of 3D domain generalization.
Furthermore, each dataset has unique characteristics, geographical locations, and classes. Ideally, we would want to have no failure class while covering different target configurations of the sensor and the environment. Therefore, an approach that leverages the advantages and characteristics of each individual dataset could provide a detection model with better generalization ability, without necessarily increasing the training time. With this as our objective, we introduce our method  for Multi-Dataset Training for 3D Object Detection (MDT3D).
We present our pipeline in Figure \ref{fig:pipe}.
We start an epoch by generating a dataset that consists of the concatenation of the same number of scans per dataset, sampled uniformly. As we load a point cloud, we apply both classical data augmentations and our novel cross-dataset augmentation. That augmented point cloud is then relabelled using our coarse labels, and fed to the chosen 3D object detection model, outputting predictions of coarse labelled boxes.

\subsection{Mixing Scenes from Different Datasets}

Instead of using a single dataset at the training time, which tends to lead to poor generalization on unseen environments and new sensor configurations, we instead took advantage of the inter-dataset point cloud distribution variance. In this way, we avoided overfitting on a single type of LiDAR distribution, and we optimized a 3D object detection model with a variety of object representations.

We define our multi-training dataset as a combination of $N$ datasets $\mathcal{D}_i, i \in \{1,...,N\}$ of size $m_i$ such that our data distribution becomes:
\begin{equation}
\mathcal{D} = \{\mathcal{D}_i\}_{i=1}^{N}
\end{equation}
with
\begin{equation}
\forall i \in \{1,...,N\}, \mathcal{D}_i = \{(\boldsymbol{p}_j, \boldsymbol{b}_j)\}_{j=1}^{m_i},
\end{equation}
where $\boldsymbol{p}_j \in \mathbb{R}^{n_j\times3}$ represents the input point clouds of $n_j$ points, and $\boldsymbol{b}_j \in \mathbb{R}^{k_j\times8}$ represents the associated $k_j$ bounding box annotations that contain the object class $c$, orientation $\theta$, localization $(x_j, y_j, z_j)$, and dimensions $(l_j, w_j, h_j)$. Furthermore, we did not use the intensity values of the 3D LiDAR points, as each dataset has a specific proprietary encoding function of the physical intensity value.

However, due to the inherent imbalance in the number of LiDAR scans found in each dataset, simply concatenating datasets may lead to an over-reliance on large datasets, such as the Waymo Open Dataset.  Therefore, in practice, for each epoch, we used the same of number of samples from each dataset to have a balanced number of LiDAR scans per dataset. 
Finally, we concatenated and shuffled the sampled pairs of scans and bounding box sets. After each epoch, we performed resampling to ensure that we would still exploit the diversity of the scans in each dataset. This should force the model to optimize its parameters from a variety of scenes between epochs while ensuring that the model would not overfit due to the prevalence of one dataset.

\subsection{Coarse Labels for a Unified Label Set}\label{sec:coarse_labels}

Each autonomous driving dataset uses its own set of labels for objects with different semantic hierarchies. For instance, the ONCE dataset separates vehicles under three categories (\textit{Car}, \textit{Truck}, \textit{Bus}) whereas the Waymo Open Dataset annotates all three of these as a single category \textit{Vehicle} from which no finer categorization can be automatically extracted. We underline these differences in Figure \ref{fig:pipe}.
The use of several datasets destabilizes the training because the datasets do not have the same class nomenclature.
Furthermore, we wanted to have a more unified set of class statistics to work with instead of having to choose those of a particular dataset so that the model would be optimized for classes of similar semantic information despite different annotations. Thus, we need a common label space and nomenclature for all datasets, so we created a set of coarse labels that regrouped the commonly annotated main moving objects within autonomous driving datasets.

To consider the various sizes of the vehicles in the datasets, we %keep 
introduced a hand-crafted separation method according to the size of the bounding box annotations.

One of the main limitations of 3D object detection models is their reliance on source dataset statistics for hyperparameter selection such as the mean size of bounding boxes for models such as PointRCNN. Thus, we separated vehicle classes into distinct categories according to their size, which will allow the model to better learn each representation while being less sensitive to dimension changes across datasets. The chosen categories are \textit{Small Vehicle}, \textit{Medium Vehicle}, \textit{Large Vehicle}, \textit{2-Wheels}.

\begin{figure}[h!]
    \includegraphics[width=0.5\textwidth]{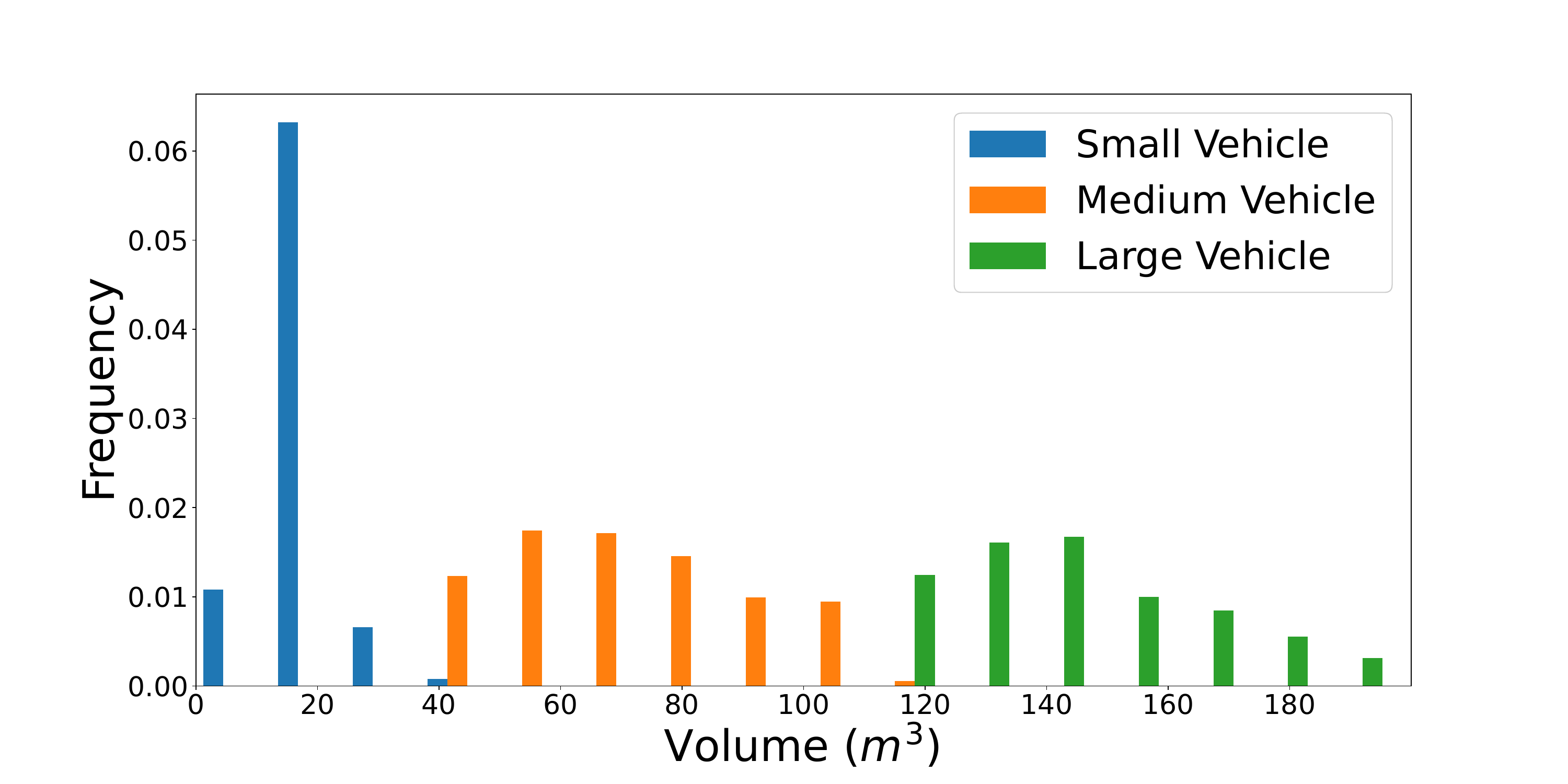}
    \caption{Histogram of the distributions of volumes of Waymo vehicles, segmented into three clusters by a K-NearestNeighbors clustering method.}
    \label{fig:histo_volume}
\end{figure}

All classes corresponding to cars will be assigned our \textit{Small Vehicle} class, the \textit{Medium Vehicle} class will correspond to small trucks, and the \textit{Large Vehicle} class to buses. We found that these represented 3 very different vehicle dimensions, which we also find in the statistics of the corresponding "fine" labels. For the Waymo dataset, which contains a single \textit{Vehicle} class with many types of vehicles classes, we applied a K-NN clustering algorithm on the volumes of the vehicles. We chose three clusters for the \textit{Vehicle} class of Waymo as we found three prevalent ranges in the histogram of ground-truth vehicle volumes for this class, as seen in Figure \ref{fig:histo_volume}. Then, we sorted these clusters according to their corresponding center and mapped these cluster centers to our three coarse labels for vehicles. This clustering step is limited to Waymo's vehicles, but could be extended to other datasets in which a coarse class contains several non-defined finer classes of different sizes. While our \textit{Pedestrian} class is the same as those in each individual datasets, we assign bicycles and motorcycles to the same coarse label \textit{2-Wheels}. We chose this label because many datasets have few annotations for bicycles and motorcycles, but their dimensions are mostly similar. Notably, KITTI does not contain any class that could be assigned to  \textit{Medium Vehicle} or \textit{Large Vehicle}, such as a \textit{Truck} or \textit{Bus} class.

\subsection{Cross-Dataset Augmentations}

To take advantage of our multiple datasets, we designed a novel data augmentation technique that mixes annotations and parts of scenes of different datasets, as an expansion of the traditional instance injection augmentation of single-dataset training\cite{Yan_Mao_Li_2018}.
These augmentations focus on learning object representations while de-emphasizing context reliance to increase generalization. We aimed to do this by injecting objects from one dataset in novel environments and sensor configurations.

We started by precomputing the mean number of instances per class for each dataset, $\bar{k}^i_c, i \in \{1,...,N\}$. Then, we chose our desired number of instances per scan $\hat{k}_c$ for each class $c$. In practice, we defined this value as the highest average number of instances from class $c$ per scan among all the training datasets. We did this to balance the number of instances across all the datasets.
Given the input point cloud $\boldsymbol{p}_j \in \mathbb{R}^{n_j\times3}$ and its corresponding bounding boxes $\boldsymbol{b}_j \in \mathbb{R}^{k_j\times8}$ from one of our $N$ datasets $\mathcal{D}_i, i \in \{1,...,N\}$,  we injected in that point cloud a number of objects per class $k^i_c$ equal to the difference between our target number of instances and the mean number of instances of that class per scan in that dataset, as follows: 
\begin{equation}
    k^i_c = \hat{k}_c - \bar{k}^i_c,
\end{equation}

To balance the injection of a given class among all datasets, we subdivided $k^i_c$ equally by the number of training datasets and injected the corresponding number of instances of this class from each training dataset in the input 3D point cloud. 

We also used data augmentations on the injected instances, namely, local rotations, $x$ and $y$ translations, translation along the distance along the axis of the instance to the ego-vehicle, rotation around the ego-vehicle, and scaling.

\section{Experiments}

\subsection{Experimental Setup}\label{sec:setup}

\textbf{Datasets.}~We built on the current autonomous driving datasets KITTI\cite{Geiger2012CVPR}, nuScenes\cite{caesar2020nuscenes}, Waymo\cite{Sun_2020_CVPR}, and ONCE\cite{once_ds}. We constructed our benchmark in a leave-one-out fashion, using three datasets as the training sets and the remaining dataset as a testing set to evaluate the generalization of our method. This was meant to validate whether our model generalizes better on unseen data than through direct transfer from a model trained on a single dataset. Due to memory constraints, when evaluating our models on Waymo, we used a fixed subsample of the evaluation set, which represented 20\% of the total scans available. 

\textbf{Evaluation Metrics.}~We used the common metric mAP to measure the accuracy of our detection models, where the average precision was calculated as the area under the precision—recall curve:
\begin{equation}
    AP = \frac{1}{40}\sum_{r=1}^{40}p(\frac{1}{r}), ~~~p = \frac{TP}{TP+FP},
\end{equation}
where $p$ is the precision, computed as the fraction of the true positive predictions over the total number of predictions, and $\frac{1}{r}$ is the recall value. While there are other dataset-specific metrics, such as the nuScenes detection score, we chose a single metric for a fair comparison of results. 

We kept the same IoU thresholds for positive detections as those in the KITTI dataset, i.e, 0.7 IoU for vehicles and 0.5 IoU for pedestrians and the \textit{2-Wheel} class.

\textbf{Implementation details.}~We used the CenterPoint detector\cite{Yin_2021_CVPR}. To demonstrate the benefits of our training paradigm independently of the model type, whether point-based or voxel-based, we also used the PointRCNN detector\cite{Shi_2019_CVPR}.

Our implementation was based on OpenPCDet\cite{openpcdet2020}. We used the same default hyperparameters for all trainings, such as a voxel size of [$0.1, 0.1, 0.2$]m for CenterPoint or radiuses of 0.2m, 1.0m, 2.0m, 4.0m, and 8.0m for the successive set aggregation layers of the PointNet++ backbone of PointRCNN. We do not optimize hyperparameters for any particular target dataset in order to avoid skewed results.

Because the training data used may not have the same size depending on the datasets used to train them, we counted the total number of LiDAR scans loaded during the training process, which is $\sim$1M, equivalent to 30 epochs for a model trained on the Waymo dataset. This way, we ensured that we were measuring the usefulness of multiple datasets rather than measuring the impact of having a larger amount of scans for training. We picked a constant batch size of 4 for all the trainings.
We utilized the Adam optimizer with a cyclical learning rate using a cosine annealer, as recommended in \cite{training_tips}, with a maximum value of $3\times10^{-3}$.

We chose the number of samples from each training dataset randomly to be equal to the minimum size among all the training datasets (see Figure \ref{fig:pipe}). 
We also used the standard data augmentation methods: scan rotation, $x/y$ translation, and $x/y$ flipping.

Furthermore, all the datasets use different heights for their LiDAR positioning and coordinate system axes. Therefore, we used manual translations along the $z$ axis and modified these coordinate systems to work within a unified canonical coordinate system based on KITTI.
For the KITTI dataset, which includes only annotations in the front-facing camera view, we removed the 3D points outside this camera view but used a stochastic flipping augmentation to have 3D scans with annotations outside this front-facing range.

\textbf{Baseline.}~ We compared our models with baselines trained on a single dataset to highlight the improved performance when the same amount of data was used but from diverse sources. Therefore, we trained both our models and the baselines using the same number of LiDAR scans, adapting the number of epochs as necessary.

\section{Results}

In this section, we show the results from our experiments to underline the increased generalization gained from our method. We also perform an ablation study to weigh the contributions of each of our modules.

\subsection{Quantitative Results}\label{ref:quantitative}

Table \ref{tab:summary_res} and Table \ref{tab:res1_1} show our results for the testing dataset that was unseen during the training, whereas the remaining datasets are used for the training. For instance, when testing our models on KITTI, we trained MDT3D using the Waymo, ONCE and nuScenes datasets. We display results for the \textit{moderate} difficulty category when testing on KITTI and remove evaluation for non-existing classes, such \textit{Medium Vehicle} and \textit{Large Vehicle}.

We first show the overall accuracy of our experiment results when transferred to an unseen target dataset. On individual datasets, MDT3D outperformed or matched the performance of the model that was trained using a single dataset, with a $+9.4$ mAP increase when testing MDT3D on ONCE with a CenterPoint model compared to simply training on the Waymo dataset. Furthermore, we compute the mean and range of test evaluation accuracies.

Despite our use of the same number of training iterations, our use of a more diverse set of 3D scans led to a stronger generalization ability. When we worked with a target domain of unknown characteristics, MDT3D acted as the best average choice and frequently surpassed single-dataset baselines.
 
Notably, we found that MDT3D acts as a strong per-class accuracy smoother.  Most classes optimized with the MDT3D training paradigm obtained the highest or second highest accuracy compared to the baselines. Additionally, we avoided the "failure class" issue, where a model trained on a single dataset would have a much worse transfer performance on some classes with very different representations in the target domain. One example is CenterPoint trained on a Waymo baseline, which reached a high transfer accuracy on the Pedestrian class but suffered when evaluating on the \textit{Small Vehicle} class when using KITTI as a target dataset. Our multi-dataset training method allowed the model to keep decent accuracy on this class while still matching the high performance of the Waymo baseline on the \textit{Pedestrian} class. However, MDT3D still fails to surpass the Oracle models on average, we believe this is due to the specific dimensions of objects in certain datasets as well as the remaining sensor gap which could be a future path of research exploration. Overall, while the accuracy of MDT3D did not match that of the Oracle model, directly trained on the target dataset, it acted as the most optimal choice of dataset on average when the test set is unknown.

\begin{table}[h]
\caption{\label{tab:summary_res}mAP results. MDT3D is trained in a leave-one-out manner.
"*" represents the case where the source and target datasets are the same (Oracle case), which we do not take into account when computing the mean and accuracy range.}
\centering
\scalebox{0.78}{
\begin{tabular}{|c||c|c|c|c|c|c|c|c|}
\hline
\multicolumn{5}{|c|}{\textbf{CenterPoint}} \\
\hline
\hline
& \multicolumn{4}{|c|}{\textbf{\textit{Target Dataset}}} \\
\hline
\textbf{\textit{Training Dataset}}& \multicolumn{1}{|c|}{\textbf{\textit{KITTI}}} & \multicolumn{1}{|c|}{\textbf{\textit{ONCE}}} & \multicolumn{1}{|c|}{\textbf{\textit{nuScenes}}} & \multicolumn{1}{|c|}{\textbf{\textit{Waymo}}} \\
 \hline
 KITTI Only & 47.4$^*$ & 9.9 & 1.6& 3.0 \\
ONCE Only & 43.0 & 60.2$^*$ &  5.8& 13.0 \\
nuScenes Only & 12.6 & 10.7& 18.4$^*$& 0.9\\
Waymo Only & 34.2 & 28.8 & 7.2& 40.2$^*$ \\
\hline
mean (min-max) & 29.9 (12.6-43.0) & 16.5 (9.9-28.8) & 4.9 (1.6-7.2) & 5.6 (0.9-13.0) \\
\hline
  MDT3D (Ours) & 41.8 & 38.2 & 11.0 & 6.4\\
\hline

\noalign{\vskip 4mm} 

\hline
\multicolumn{5}{|c|}{\textbf{PointRCNN}} \\
\hline
\hline
& \multicolumn{4}{|c|}{\textbf{\textit{Target Dataset}}} \\
\hline
\textbf{\textit{Training Dataset}} & \multicolumn{1}{|c|}{\textbf{\textit{KITTI}}} & \multicolumn{1}{|c|}{\textbf{\textit{ONCE}}} & \multicolumn{1}{|c|}{\textbf{\textit{nuScenes}}} & \multicolumn{1}{|c|}{\textbf{\textit{Waymo}}} \\
\hline
KITTI Only & 60.6$^*$ & 10.9 &  4.8& 3.0 \\
ONCE Only & 48.9 & 36.3$^*$&  7.9 & 7.8 \\
nuScenes Only & 23.0 & 10.2& 10.5$^*$& 5.9\\
Waymo Only & 32.3 & 16.0 & 7.6& 20.3$^*$ \\
\hline
mean (min-max) &  34.7 (23.0-48.9) & 12.4 (10.2-16.0) & 6.8 (4.8-7.9)&5.6 (3.0-7.8) \\
\hline
MDT3D (Ours) & 49.2 & 18.9 & 7.5&8.4 \\
  \hline
\end{tabular}}
\end{table}

\begin{table*}[t]
\vspace{5pt}
\caption{\label{tab:res1_1}Detailed Per-Class AP Results for the CenterPoint model. "*" represents the oracle case, which we do not take into account when highlighting the best result in bold.}
\centering
\scalebox{1}{
\begin{tabular}{|c||c|c|c||c|c|c|c|c|c||c|}
\hline
\multicolumn{11}{|c|}{\textbf{CenterPoint}} \\
\hline
\hline
& \multicolumn{10}{|c|}{\textbf{\textit{Target Dataset}}} \\
\hline
 & \multicolumn{4}{|c|}{\textbf{\textit{KITTI}}} & \multicolumn{6}{|c|}{\textbf{\textit{ONCE}}} \\
\hline
 \textbf{\textit{Training Dataset}} & \textbf{\textit{SmallV}} & \textbf{\textit{2-Wheels}} & \textbf{\textit{Ped}} & \textbf{\textit{mAP}} & \textbf{\textit{SmallV}} & \textbf{\textit{MedV}} & \textbf{\textit{LargeV}} & \textbf{\textit{2-Wheels}} & \textbf{\textit{Ped}} & \textbf{\textit{mAP}} \\
  \hline
  KITTI only & 68.2$^*$ & 42.4$^*$  & 31.5$^*$  & 47.4$^*$  & 23.2 & 0 & 0 & 2.4 & 4.2 &  9.9\\
  ONCE only & \textbf{55.5} & 40.1 & 33.5 & \textbf{43.0} & 83.0$^*$  & 36.8$^*$  & 72.2$^*$  & 74.8$^*$  & 34.1$^*$  & 60.2$^*$  \\
  nuScenes Only & 7.8 & 1.7 & 28.2 & 12.6 & 18.0 & 4.6 & 17.8 & 2.7 & 10.4 & 10.7\\
  Waymo only & 11.6 & \textbf{43.6} & 47.4 & 34.2 & 46.3 & 4.4 & 25.6 & 23.9 & \textbf{43.6} & 28.8 \\
  \hline
  MDT3D (Ours) & 35.2 & 42.2 & \textbf{48.0} & 41.8 & \textbf{49.0} & \textbf{5.9} & \textbf{39.7} & \textbf{26.8} & 40.0 & \textbf{38.2} \\ 
  \hline
\end{tabular}}
\end{table*}

\subsection{Ablation Study}

 While the use of a unified label set is necessary for training, we studied the effect of the components of our methods on a CenterPoint model using ONCE as a target in Table \ref{tab:ablation}. Interestingly, our data sampling method seemed to lower our target performance ($-1.4$ mAP on the ONCE dataset). This is due to the imbalance in datasets used for training. Models on Waymo tend to transfer best on the ONCE dataset and in this case Waymo represented the majority dataset when we simply concatenated them without data sampling. However, removing the data sampling component led to a much larger drop in source dataset accuracy ($-8.2$ mAP on the KITTI dataset). Therefore, in some cases no sampling will lead to a drop in accuracy if transferring to a domain closer to that of KITTI. Adding our cross-dataset instance injection added a substantial increase to the accuracy, suggesting the insertion of objects in novel contexts may help improve the generalization of 3D detectors.
\begin{table}[h!]
\caption{Ablation Study of each component}
\scalebox{1}{
    \centering
    \begin{tabular}{|c|c|c|c|}
    \hline
         \textbf{\textit{Data Sampling}} & \textbf{\textit{Cross-Dataset Instance Injection}} & \textbf{\textit{ONCE mAP}} \\%& Waymo mAP \\
        \hline
         &  & 30.0\\% & X\\ 
%11.8+1.2+7+25.7+13.6
         \checkmark &  & 28.6 \\%& 8.9\\ 
         \checkmark & \checkmark  & 38.2\\% & 11.0\\ 
         \hline
    \end{tabular}}
    
    \label{tab:ablation}
\end{table}
\vspace{-6pt}
\subsection{Iteration Scaling}

\begin{table}
\caption{Evolution of mAP on the ONCE dataset.}
\scalebox{0.9}{
    \centering
    \begin{tabular}{|c|c|c|c|c|c|c|c|}
    \hline
        & \multicolumn{7}{|c|}{\textbf{Number of Iterations}} \\
        \hline
         \textbf{\textit{Method}} & \textbf{\textit{0.5M}} & \textbf{\textit{1M}} & \textbf{\textit{1.5M}} & \textbf{\textit{2M}} & \textbf{\textit{2.5M}} & \textbf{\textit{3M}} & \textbf{\textit{3.5M}}\\
        \hline
         Waymo Only &  26.0 & 28.8 & 29.1 & 29.3 & 29.7& 29.7& 29.8\\
         MDT3D (Ours) &  23.4 & 28.6 & 29.9 & 30.2 & 31.1& 31.4& 32.0\\
         \hline
    \end{tabular}}
    \label{tab:num_epochs}
\end{table}

To find whether MDT3D benefits from increasing the number of iterations, we trained a CenterPoint model using our MDT3D method and compared it with a CenterPoint model trained on the Waymo dataset, testing on the ONCE dataset. Our MDT3D was trained without the cross-dataset instance injection in order to underline how simply increasing the number of iterations allows MDT3D to surpass other baselines. We compared three settings, starting by using the same number of iterations as in the experiments of \ref{ref:quantitative}. We also sub-trained using half this number of iterations, which led to slight underfitting. Finally, we increased the number of iterations by half, which traditionally leads to overfitting and should have slightly diminishing returns in transfer accuracy, particularly on less diverse training sets. We show the evolution of the mAP for both methods in Table \ref{tab:num_epochs}. At half the iterations, MDT3D reached a mAP of $23.4$ while the Waymo baseline had a $26.0$ mAP.  With this lower number of iterations, MDT3D was hampered by the higher diversity in its training data. However, as we increased the number of iterations, MDT3D ended up outperforming the baseline trained on Waymo ($32.0$ mAP for MDT3D with $\sim3.5$M iterations and $29.8$ for Waymo). Therefore, we believe that MDT3D benefits from a higher number of iterations when trying to increase its generalization ability and that its multi-dataset diversity allows the model to scale with the number of iterations with weaker overfitting than single-dataset models.

\section{Conclusion}

In this paper, we have proposed a new multi-source training paradigm for 3D object detection in autonomous driving scenarios. This approach allows us to leverage the variety of scenes and sensor configurations present in various LiDAR datasets.
We find that using the same amount of training iterations, but from diverse sources of data can lead to models that generalize better to new unseen datasets, without having to manually engineer hyperparameters. 
This training is enabled by working on a coarse set of labels which allows for a stable training. Finally, using multi-dataset information allows us to use a cross-dataset data augmentation in the form of cross-dataset object injection, further enhancing the generalization of trained models. 
 In the future, we believe that integrating the sampling as a jointly learned component of the network could improve the generalization capability of the neural network by using clusters of extracted features and geometric information from which to sample.
 
\section{Acknowledgements}
This project has received funding from the European Union’s Horizon 2020 research and innovation program under grant agreement No. 951947 (5GMed).
\vspace{-1pt}
\bibliographystyle{ieeetr}
\bibliography{ds, multids, generalization, obj_det,
              domain_adaptation, other}

\end{document}